\documentclass{article}

\usepackage{microtype}
\usepackage{graphicx}
\usepackage{subcaption}
\usepackage{booktabs} 

\usepackage{hyperref}

\usepackage[preprint]{icml2026}

\usepackage{amsmath}
\usepackage{amssymb}
\usepackage{mathtools}
\usepackage{amsthm}

\usepackage[capitalize,noabbrev]{cleveref}

\theoremstyle{plain}

\theoremstyle{definition}

\theoremstyle{remark}

\usepackage[textsize=tiny]{todonotes}

\icmltitlerunning{Slumbering to Precision: Enhancing Artificial Neural Network Calibration Through Sleep-like Processes}

\begin{document}

\twocolumn[
  \icmltitle{Slumbering to Precision: Enhancing Artificial Neural Network Calibration Through Sleep-like Processes}


\begin{icmlauthorlist}
  \icmlauthor{Jean Erik Delanois}{ucsdcs,ucsdmed}
  \icmlauthor{Aditya Ahuja}{ucsdcs,ucsdmed}
  \icmlauthor{Giri P. Krishnan}{gatech}
  \icmlauthor{Maxim Bazhenov}{ucsdmed}
\end{icmlauthorlist}

\icmlaffiliation{ucsdcs}{Department of Computer Science \& Engineering, University of California, San Diego, La Jolla, California, USA}
\icmlaffiliation{gatech}{ARTISAN, Georgia Institute of Technology, Atlanta, Georgia, USA}
\icmlaffiliation{ucsdmed}{Department of Medicine, University of California, San Diego, La Jolla, California, USA}

\icmlcorrespondingauthor{Maxim Bazhenov}{mbazhenov@ucsd.edu}

\icmlkeywords{Machine learning, Neural network calibration, Uncertainty estimation,
Sleep replay}

  \vskip 0.3in
]



\printAffiliationsAndNotice{}  

\begin{abstract}
Artificial neural networks are often overconfident, undermining trust because their predicted probabilities do not match actual accuracy. Inspired by biological sleep and the role of spontaneous replay in memory and learning, we introduce Sleep Replay Consolidation (SRC), a novel calibration approach. SRC is a post-training, sleep-like phase that selectively replays internal representations to update network weights and improve calibration without supervised retraining. Across multiple experiments, SRC is competitive with and complementary to standard approaches such as temperature scaling. Combining SRC with temperature scaling achieves the best Brier score and entropy trade-offs for AlexNet and VGG19. These results show that SRC provides a fundamentally novel approach to improving neural network calibration. SRC-based calibration offers a practical path toward more trustworthy confidence estimates and narrows the gap between human-like uncertainty handling and modern deep networks.
\end{abstract}

\section{Introduction}

Artificial neural networks (ANNs) are increasingly integral to high-stakes applications such as healthcare diagnostics, autonomous driving, and financial forecasting. Consequently, it is essential for human operators to both understand and trust the outputs generated by these systems. However, current deep ANNs often struggle to reliably estimate the confidence of their predictions \cite{abdar2021review,gawlikowski2023survey,guo2017calibration}. A key component of reliability across various tasks is calibration — the alignment between the model’s predicted confidence (the probability assigned to a prediction) and the actual likelihood of being correct. Poorly calibrated models, where confidence fails to reflect true accuracy, can lead to critical errors: users may place undue trust in inaccurate predictions or disregard highly probable ones, with potentially serious consequences.

Specifically, modern ANNs often exhibit overconfidence in their predictions, meaning the predicted probabilities are systematically higher than the true likelihood of being correct \cite{guo2017calibration};
this causes ANNs to report high confidence even on incorrect examples. Poor ANN calibration
contrasts with human decision-making where individuals can 
effectively balance certainty and uncertainty.


Current post-hoc approaches for improving calibration, such as Temperature Scaling (TS) \cite{guo2017calibration} and Confidence-Based Temperature (CBT) \cite{frenkel2022network}, typically rely on simple renormalization of output-layer activations, which uniformly smooths the output distribution and reduces predicted confidence. More expressive post-hoc schemes, including Dirichlet calibration \cite{kull2019beyond} and related multiclass reductions \cite{gupta2022top,verma2022calibrated}, extend this framework but remain output-level transformations that do not alter the underlying model parameters. In contrast, other calibration strategies - such as label smoothing \cite{muller2019does}, focal loss \cite{mukhoti2020calibrating}, and weight regularization \cite{guo2017calibration} - require explicit retraining with modified objectives. As a result, there remains a gap in post-hoc methods that directly modify network weights to genuinely reshape model confidence. 

Multiple lines of evidence indicate that sleep plays a critical role in shaping human judgment and confidence. By strengthening relevant memory traces and reducing representational noise, sleep helps align subjective confidence with actual performance \cite{Paller2020,Payne2008,Drosopoulos2005}. Through memory reorganization, insight generation, and hippocampal–neocortical replay, sleep sharpens distinctions between strong and weak memories, improving confidence accuracy and correcting misconceptions \cite{Wagner2004,Diekelmann2010,Rasch2007,Whitmore2022}. In contrast, sleep deprivation disrupts this calibration: individuals become overconfident, underestimate errors, and fail to update beliefs \cite{Killgore2010,McKenna2007}. Accordingly, human studies show that sleep loss impairs confidence–accuracy alignment, while normal or recovery sleep restores accurate confidence judgments \cite{Baranski1994Calibration,Blagrove2000Confidence,Baranski2007Fatigue,Sundelin2024Metacognition}.


Sleep-dependent memory consolidation is mediated by spontaneous reactivation (replay) of memory traces, leading to synaptic modification \cite{wei2016synaptic,gonzalez2020can,golden2020sleep}. 
Prior work demonstrated several benefits of sleep-like processing - Sleep Replay Consolidation (SRC) - in artificial neural networks, including reduced catastrophic forgetting in continual learning \cite{Tadros_NC2022}, improved generalization \cite{Delanois_ICMLA23}, and increased accuracy in data-constrained settings \cite{Bazhenov_AAAI24}. 

In contrast to prior studies, we show that SRC induces a qualitatively new capability: post-hoc calibration of already-trained networks through principled weight updates, aligning predictive confidence with accuracy without any additional training, fine-tuning, or supervision. We further demonstrate that this sleep-based approach scales from small feedforward models to deep, widely used CNN architectures, including CIFAR-100 and ImageNet. More broadly, our results introduce a new class of post-hoc calibration methods that operate by modifying network weights rather than applying simple output-level transformations.

Integrating sleep-like stages into ANNs offers a practical approach to model calibration that bridges post-hoc and retraining-based methods. SRC operates on fully trained networks but improves calibration through direct weight modification, avoiding both simplistic output remapping and the high computational cost of retraining. By enabling efficient offline weight adaptation, SRC enhances the reliability of pretrained models while preserving deployment efficiency, providing a path toward more trustworthy AI systems.

\paragraph{Main contributions:}

\begin{itemize}
  \setlength{\itemsep}{2pt}
  \setlength{\parskip}{0pt}
  \setlength{\parsep}{0pt}
\item We propose a new approach to improving ANN calibration: unsupervised Sleep Replay Consolidation (SRC). SRC matches or outperforms existing post-hoc techniques and can rival more resource-intensive retraining approaches.
\item SRC is a unique post-hoc method capable of altering model weights after training without labels and improve calibration.
\item Our analysis suggests that SRC improves calibration by augmenting features and increasing sparsity in internal representations. These effects are complementary to the mechanisms of temperature scaling and label smoothing - common calibration methods that typically do not induce such representational changes.

\item SRC is compatible with other post-hoc methods, and combining it with existing approaches lead to state-of-the-art calibration performance.

\end{itemize}

\section{Methods}


In this work, we benchmark ResNet, GoogLeNet, AlexNet, and VGG on ImageNet and CIFAR-100 \cite{he2016deep, szegedy2015going,krizhevsky2012imagenet, simonyan2014very, deng2009imagenet, krizhevsky2009learning}.


\subsection{Model Architecture and Training Paradigm}

\paragraph{CIFAR-100 Models:}
CIFAR-100 experiments used a ResNet-152 backbone with a fully connected feedforward head (1024, 512, 100). The model was trained for 100 epochs with cross-entropy loss using SGD (learning rate 0.001, momentum 0.9, L2 regularization 0.001, batch size 32), with a 10\% learning rate decay every 50 epochs. Data augmentation included horizontal flips (50\%), sharpness enhancement (10\%), and color jitter (40\%).

\paragraph{ImageNet Models:}
For ImageNet, standard pretrained networks were used. Models with an added multi-layer FF head included two 2048-unit ReLU hidden layers and a 1000-unit output layer. The backbone was frozen, and the FF head was trained for five epochs using cross-entropy loss with SGD (learning rate 0.1, batch size 256) and dropout of 0.2 in the hidden layers.

\paragraph{Simulation Paradigm:}
Each experiment used 10 trials with different random seeds. Pretrained or fully trained models served as the Baseline. SRC was then applied to the feedforward head (Baseline + SRC), using convolutional output statistics as input. SRC hyperparameters were tuned via a genetic algorithm on the validation set. For ImageNet, the validation set was split evenly for tuning and testing due to unavailable test labels, and results were reported on the held-out test set.


For ResNet-152 on CIFAR-100, retraining with label smoothing (Baseline + LS) yielded an optimal smoothing value of 0.05 from a sweep over 0.05–0.2. Retraining with Focal Loss (Baseline + Focal) identified optimal parameters $\alpha = 1$ and $\gamma = 1$ from sweeps over $\alpha \in [0.1, 1]$ and $\gamma \in [1, 4]$.

Temperature scaling (Baseline + TS, Baseline + SRC + TS) was optimized using the L-BFGS algorithm (learning rate 0.1, up to 400 iterations) on the same validation set used for SRC tuning, with results reported on the test set. If L-BFGS produced worse calibration than the baseline, those results were retained; effectively, when no improvement was found, the original model was used (equivalent to temperature = 1).

\begin{table*}[t!]
\centering
\resizebox{0.7\textwidth}{!}{%
\begin{tabular}{@{}cccccc@{}}
\toprule
 & Accuracy & ECE & NLL & Brier & Entropy \\

\midrule

\textbf{ResNet 152 on CIFAR 100} \\
Baseline & \textbf{84.9} & 0.062184 & 0.601612 & 9.369636 & 0.457585 \\
Baseline + SRC & 84.8 & \textcolor{black}{\textbf{0.012978}} & 0.570327 & \textcolor{black}{\textbf{4.33612}} & \textcolor{black}{\textbf{1.015678}} \\
Baseline + TS & \textbf{84.9} & 0.016904 & \textbf{0.562992} & 5.478358 & 0.838229 \\
Baseline + SRC + TS & 84.8 & 0.014159 & 0.565073 & 4.940667 & 0.857810 \\
\\
Baseline + LS & \textbf{85.2\textsuperscript{\textdagger}} & 0.026636 & 0.617318 & \textbf{1.624467\textsuperscript{\textdagger}} & \textbf{1.162224\textsuperscript{\textdagger}} \\
Baseline + Focal  & 85.0 & \textbf{0.010139\textsuperscript{\textdagger}} & \textbf{0.542586\textsuperscript{\textdagger}} & 6.243289 & 0.837582 \\

\\
\textbf{AlexNet on ImageNet} \\
Baseline & \textbf{55.7} & 0.016945 & 1.938792 & 12.877128 & 2.553596 \\
Baseline + SRC & 55.6 & \textcolor{black}{\textbf{0.012842}} & 1.944034 & 11.487936 & 2.716625 \\
Baseline + TS & \textbf{55.7} & 0.016686 & \textbf{1.933153} & 11.203075 & 2.799049 \\
Baseline + SRC + TS & 55.6 & 0.018179 & 1.943310 & \textcolor{black}{\textbf{10.8851737}} & \textcolor{black}{\textbf{2.8143639}} \\
\\
\textbf{VGG19 on ImageNet}\\
Baseline & \textbf{71.6} & 0.028622 & 1.136530 & 10.980175 & 1.394414 \\
Baseline + SRC & 71.5 & \textcolor{black}{\textbf{0.0166583}} & 1.129583 & 9.298542 & 1.583745 \\
Baseline + TS & \textbf{71.6} & 0.017769 & \textbf{1.127943} & 9.090689 & 1.612490 \\
Baseline + SRC + TS & 71.5 & 0.017128 & 1.129311 & \textcolor{black}{\textbf{9.0486836}} & \textcolor{black}{\textbf{1.6174512}} \\
\\
\textbf{ResNet 50 on ImageNet} \\
Baseline & 75.3 & 0.041667 & 0.984836 & 6.107409 & 1.106435 \\
Baseline + SRC & \textcolor{black}{\textbf{75.3}} & 0.041764 & 0.985621 & 6.035420 & 1.104285 \\
Baseline + TS & 75.3 & \textbf{0.0195233} & \textbf{0.970407} & 4.750875 & 1.390268 \\
Baseline + SRC + TS & \textcolor{black}{\textbf{75.3}} & 0.019830 & 0.970895 & \textcolor{black}{\textbf{4.682062}} & \textcolor{black}{\textbf{1.3912049}} \\
\\
\textbf{ResNet 152 on ImageNet} \\
Baseline & \textbf{77.7} & 0.054090 & 0.896585 & 6.154923 & 0.861534 \\
Baseline + SRC & 77.6 & 0.053647 & 0.903076 & 6.514544 & 0.869440 \\
Baseline + TS & \textbf{77.7} & \textbf{0.019601} & \textbf{0.8659595} & \textbf{4.1824167} & 1.241993 \\
Baseline + SRC + TS & 77.6 & 0.020916 & 0.872437 & 4.433809 & \textcolor{black}{\textbf{1.250647}} \\
\\
\textbf{ResNet 152 FF on ImageNet} \\
Baseline & \textbf{76.6} & 0.078514 & 0.993487 & 7.911834 & 0.778780 \\
Baseline + SRC & 75.9 & \textcolor{black}{\textbf{0.0202338}} & 0.965635 & 4.796003 & \textcolor{black}{\textbf{1.4393607}} \\
Baseline + TS & \textbf{76.6} & 0.022432 & \textbf{0.926918} & \textbf{4.566641} & 1.330852 \\
Baseline + SRC + TS & 75.9 & 0.021258 & 0.965318 & 4.886078 & 1.410158 \\
\\
\textbf{GoogLeNet on ImageNet} \\
Baseline & \textbf{69.7} & 0.062274 & 1.283732 & \textbf{1.619943} & 2.636590 \\
Baseline + SRC & 60.1 & 0.015086 & 1.962518 & 9.750454 & \textcolor{black}{\textbf{2.9477475}} \\
Baseline + TS & \textbf{69.7} & 0.019073 & \textbf{1.2386713} & 2.271732 & 1.810560 \\
Baseline + SRC + TS & 60.1 & \textcolor{black}{\textbf{0.014054}} & 1.961691 & 10.103846 & 2.843713 \\
\\
\textbf{GoogLeNet FF on ImageNet} \\
Baseline & \textbf{68.3} & 0.041308 & 1.286850 & 5.089959 & 1.555107 \\
Baseline + SRC & 68.2 & 0.015631 & 1.280730 & \textcolor{black}{\textbf{4.026876}} & \textcolor{black}{\textbf{1.942972}} \\
Baseline + TS & \textbf{68.3} & \textbf{0.012558} & \textbf{1.2770282} & 4.175807 & 1.864093 \\
Baseline + SRC + TS & 68.2 & 0.014230 & 1.278946 & 4.184295 & 1.868947 \\
\bottomrule
\end{tabular}%
}
\caption{Accuracy and calibration metrics averaged over 10 trials (STD in Appendix B). Bold values indicate the best-performing post-hoc method. Bold values marked with (\textsuperscript{\textdagger}) indicate methods that achieved the best score but required retraining.}
\label{tab:main}
\vspace{-1mm}
\end{table*}

\subsection{Sleep Replay Consolidation (SRC) algorithm}\label{sec:SRC}


We implement SRC following \cite{Tadros_NC2022}, where it mitigated catastrophic forgetting in class-incremental learning. In our setup, SRC is applied only to the feedforward head (see Appendix). SRC maps the ANN to a spiking neural network (SNN) with identical architecture, as in \cite{diehl2015fast,Tadros_NC2022}. Activations are replaced with a Heaviside function, and weights are scaled by the maximum layer-wise activation observed during prior training to ensure stable firing activity \cite{diehl2015fast}. The network then undergoes successive forward passes driven by stochastic spike trains. For each input vector, spike probabilities are drawn from a Poisson distribution with mean rates proportional to feature-wise average intensities across previously seen training data, such that higher-mean features produce more active input neurons. After each forward pass, synaptic weights are updated via an unsupervised Hebbian rule: weights increase when pre- and post-synaptic neurons co-activate and decrease when post-synaptic activity occurs without pre-synaptic activation. After multiple offline iterations, weights are rescaled back, the original activation function is restored, and the network is returned to standard ANN operation. SRC hyperparameters were optimized separately for each model using a genetic algorithm to maximize validation accuracy.

To prevent overfitting, tuning and evaluation data were strictly separated. For ImageNet, the validation set was split evenly, with one half used to tune SRC and temperature scaling (TS) and the other reserved for evaluation. For CIFAR-100, SRC and TS were tuned on a held-out set, and all metrics were reported on an unseen test set, ensuring no method was evaluated on data used for hyperparameter selection.

\begin{figure*}[ht]
\begin{center}
\includegraphics[width=0.8\columnwidth]{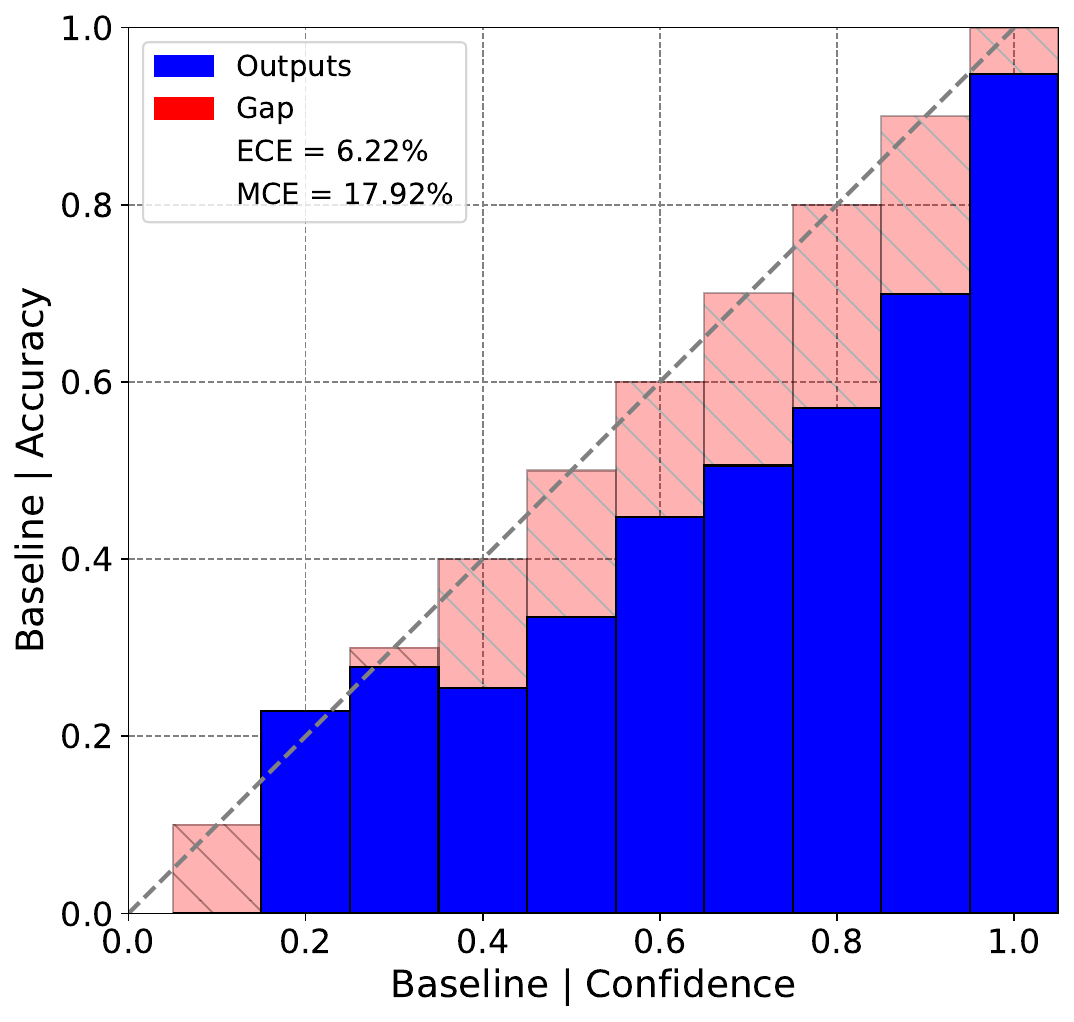}
\includegraphics[width=0.8\columnwidth]{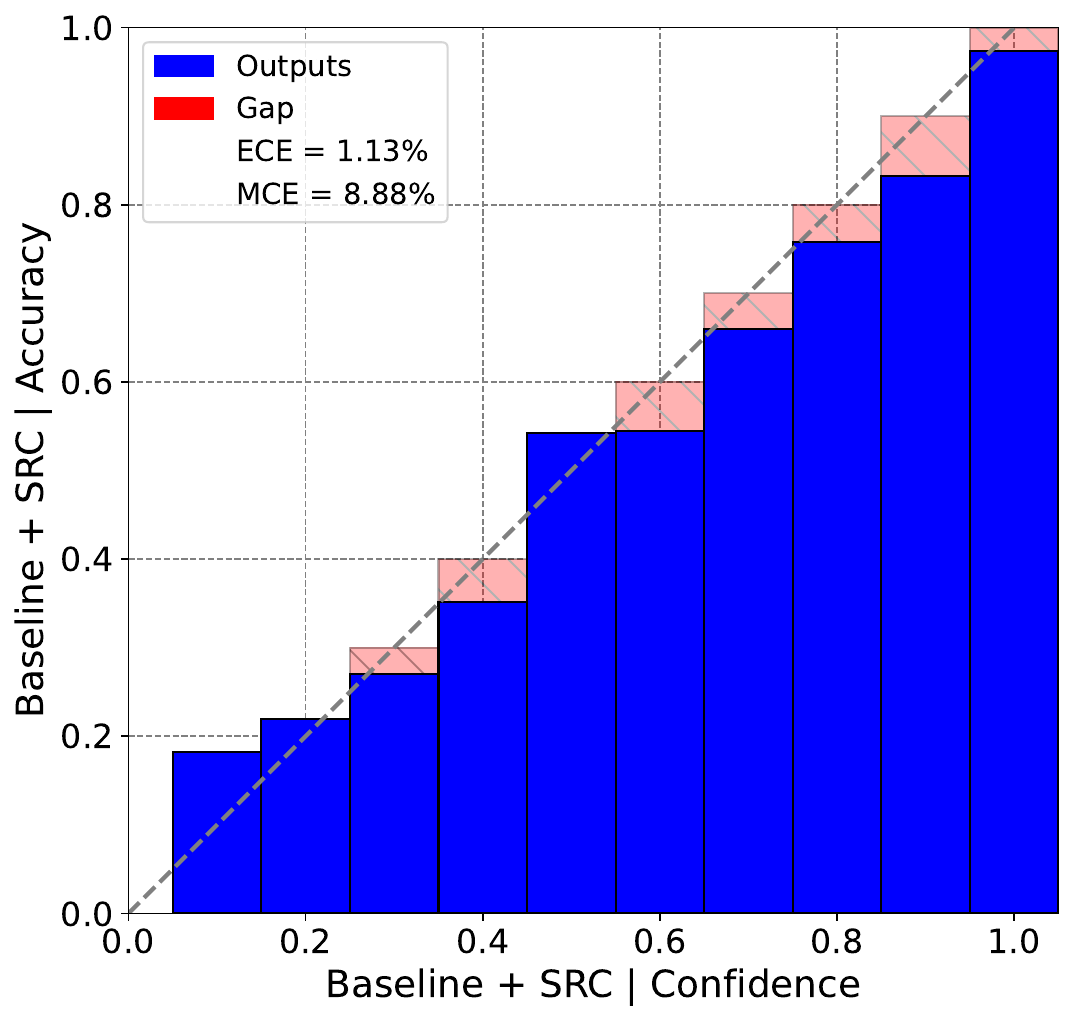}
\caption{Reliability diagrams showing improvement in ECE metric as a result of SRC (Right) after initial Baseline training (Left) for the ResNet-152 model trained on CIFAR-100. Confidence more accurately reflects accuracy (blue bins closer to diagonal). Red bars indicate difference between ideal and actual accuracy. Test confidence and accuracy are shown.}
\label{fig:cifar100-ResNet152-ece}
\end{center}
\end{figure*}

\section{Results}
\subsection{Impact of SRC on Calibration Metrics}

To evaluate the unsupervised post-hoc Sleep Replay Consolidation (SRC) algorithm, we conducted comparative experiments across methods, models, and datasets, alongside analyses of confidence and feature representations. Models used standard naive or pretrained CNN backbones with either original or fine-tuned multi-layer feedforward (FF) heads. We tested three settings: (1) ImageNet-pretrained models evaluated on ImageNet, (2) ImageNet-pretrained backbones with retrained FF heads evaluated on ImageNet, and (3) end-to-end trained CNNs with multi-layer FF heads evaluated on CIFAR-100. Our primary analysis focused on CIFAR-100, with results validated on ImageNet.

SRC consistently improved calibration across metrics (see mean over 10 trials in Table \ref{tab:main}, standard deviations in Appendix). On CIFAR-100, SRC was compared against post-hoc Temperature Scaling (TS) and retraining with Label Smoothing (LS) or Focal Loss; on ImageNet, only post-hoc methods were considered due to computational constraints. The largest and most consistent gains occurred in models with multi-layer FF heads, indicating that SRC primarily improves calibration via feature-level adaptation rather than decision-layer changes. Models with single output layers (e.g., ResNet on ImageNet) showed minimal improvement without degradation. Notably, adding a multi-layer FF head preserved accuracy while enabling SRC to substantially improve calibration; for example, in ResNet-152 with a custom FF head, ECE dropped from 
0.0785 to 0.0202.



\paragraph{Accuracy:}
Importantly, SRC does not generally degrade accuracy. Across models, overall accuracy remains largely unchanged after SRC (Table \ref{tab:main}), with only one exception (GoogLeNet). This drop was eliminated when a custom multi-layer head is prepended (GoogLeNetFF vs. GoogLeNet; Table \ref{tab:main}), highlighting that SRC is most effective when acting on feature representations. While some applications may prioritize calibration over raw accuracy, preserving predictive performance is clearly preferable. Overall, our results show that SRC can enhance calibration through synaptic modification while maintaining accuracy - without storing or replaying additional training examples.

\paragraph{Expected Calibration Error (ECE):}
Expected Calibration Error (ECE) \cite{naeini2015obtaining} directly quantifies the mismatch between confidence and accuracy (Appendix A). After applying SRC, ECE consistently decreased across all models (Table \ref{tab:main}), indicating improved confidence–accuracy alignment. Notably, the best post-hoc ECE was achieved by SRC alone in 4 of 8 models, and by SRC+TS in one model.
This demonstrates the effectiveness of SRC and its ability to work synergistically with post-hoc methods such as TS. Importantly, SRC alone outperformed some retraining approaches (e.g., ResNet-152 on CIFAR-100, SRC vs. LS; Table \ref{tab:main}).


SRC’s effect on ECE was modest in models with a single-layer feedforward head (e.g., ResNet-152 on ImageNet; Table \ref{tab:main}), but became substantially stronger after adding multiple FF layers (ResNet-152FF).
This again suggests that SRC is particularly effective in deeper FF heads, consistent with its mechanism of refining feature representations to enhance calibration.

SRC’s impact on ECE is illustrated in Figure \ref{fig:cifar100-ResNet152-ece} for ResNet-152 on CIFAR-100. Samples are binned by predicted confidence (x-axis), with bar heights indicating empirical accuracy. Figure \ref{fig:cifar100-ResNet152-ece} (left) shows poor baseline calibration, with large deviations from the diagonal indicating misaligned confidence.
After SRC, calibration improves markedly (Figure \ref{fig:cifar100-ResNet152-ece}, right), as bars align closely with the diagonal, reflecting a substantially improved match between confidence and accuracy.


Unlike TS, which only remaps the output distribution, SRC modifies synaptic structure and feature representations to improve calibration. While TS masks poor calibration, SRC reshapes the model’s internal notion of confidence.





\paragraph{Negative Log Likelihood (NLL):}



NLL is a standard loss function often used as a proxy for calibration \cite{guo2017calibration} (Appendix A). However, as shown in Table \ref{tab:main}, NLL varied much less across architectures than other calibration metrics. For instance, on CIFAR-100, SRC and TS reduced ECE by 79.2\% and 72.8\%, respectively, yet improved NLL by only 5.1\% and 6.4\%.

Notably, some retraining-based approaches slightly increased NLL despite improving other calibration measures (e.g., ResNet-152 on CIFAR-100, Baseline + LS; Table \ref{tab:main}). Given that retraining methods are among the most invasive and are known to improve calibration, LS’s failure to reduce NLL suggests that NLL does not fully capture calibration.

In contrast, TS consistently achieved the lowest NLL across post-hoc settings, albeit with modest gains. This aligns with its mechanism: NLL reflects the log-probability assigned to the correct class, whereas other metrics capture binned accuracy or distribution-level properties. By smoothing overconfident predictions - especially for misclassified samples - TS reallocates probability mass toward the correct class, directly reducing NLL.

The fact that TS is the only method to consistently improve NLL underscores that it targets a specific mode of miscalibration that NLL is sensitive to, whereas other methods (like SRC) improve calibration through different modes. This supports the broader conclusion that multiple post-hoc methods are needed to comprehensively address calibration across metrics.







\paragraph{Brier Score:}

The application of SRC consistently yielded strong results in terms of the Brier score (Table \ref{tab:main}), which quantifies the mean squared difference between predicted probabilities and actual outcomes (Appendix A). Furthermore, SRC demonstrated synergistic effects when combined with TS, frequently resulting in optimal post-hoc Brier (5 models out of 8).



\enlargethispage{\baselineskip}
\paragraph{Entropy:}
Entropy as measure of uncertainty or randomness (Appendix A), serves as a valuable proxy for network calibration \cite{mukhoti2020calibrating}. Low entropy indicates that the network assigns most of the probability mass to a single output class, often reflecting overconfidence in its predictions and resulting in poor calibration. In contrast, higher entropy reflects a more balanced probability distribution across multiple classes, suggesting reduced overconfidence and improved calibration. We found the incorporation of SRC always increased entropy and yielded the highest post-hoc entropy for all models tested. SRC even increased entropy more than retraining the model with Focal Loss (ResNet 152 on CIFAR 100 SRC vs Focal Table \ref{tab:main}), signifying the significant improvement in network calibration.

\paragraph{Summary:}

SRC reliably sharpens probabilistic calibration while leaving classification accuracy intact.  It \textit{matches, surpasses, or complements} TS on pure calibration metrics (ECE) in 5 out of 8 Models. SRC excels on networks equipped with deep FF heads, and pairs seamlessly with TS to yield the best holistic error (Brier and Entropy).  Given that SRC is unsupervised and parameter‑light, it offers a practical drop‑in upgrade for legacy models at scale. 

In practice, SRC is preferable over TS when stronger calibration improvements are required and an offline replay phase is feasible. Because SRC is applied offline, accuracy can always be validated and the original model or TS retained if performance degrades.

\begin{figure*}[ht]
\begin{center}
\includegraphics[width=0.9\columnwidth]{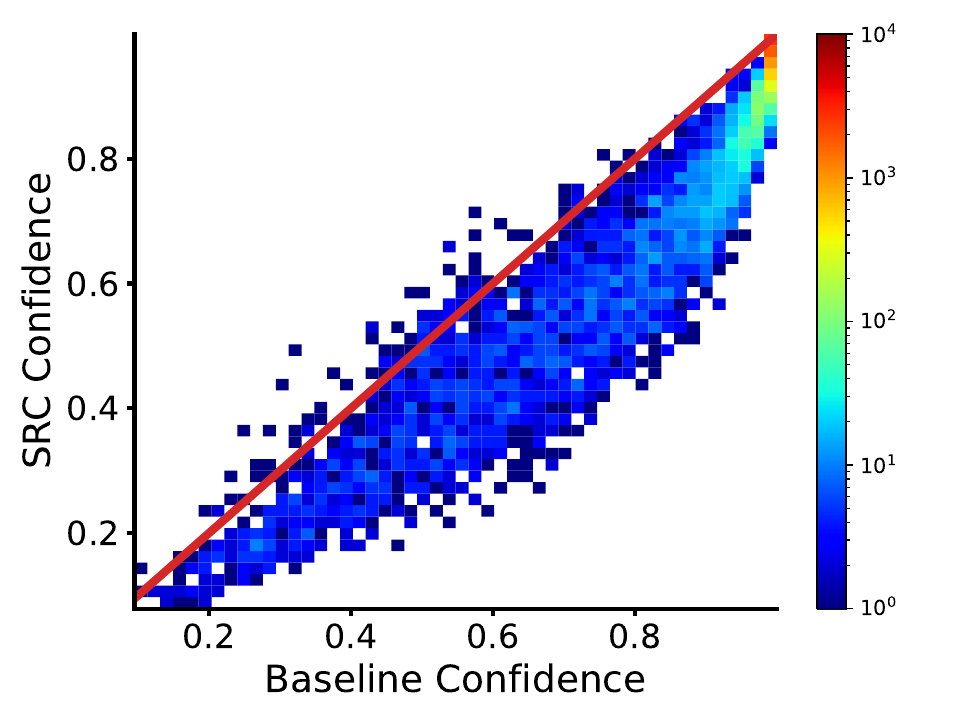}
\includegraphics[width=0.9\columnwidth]{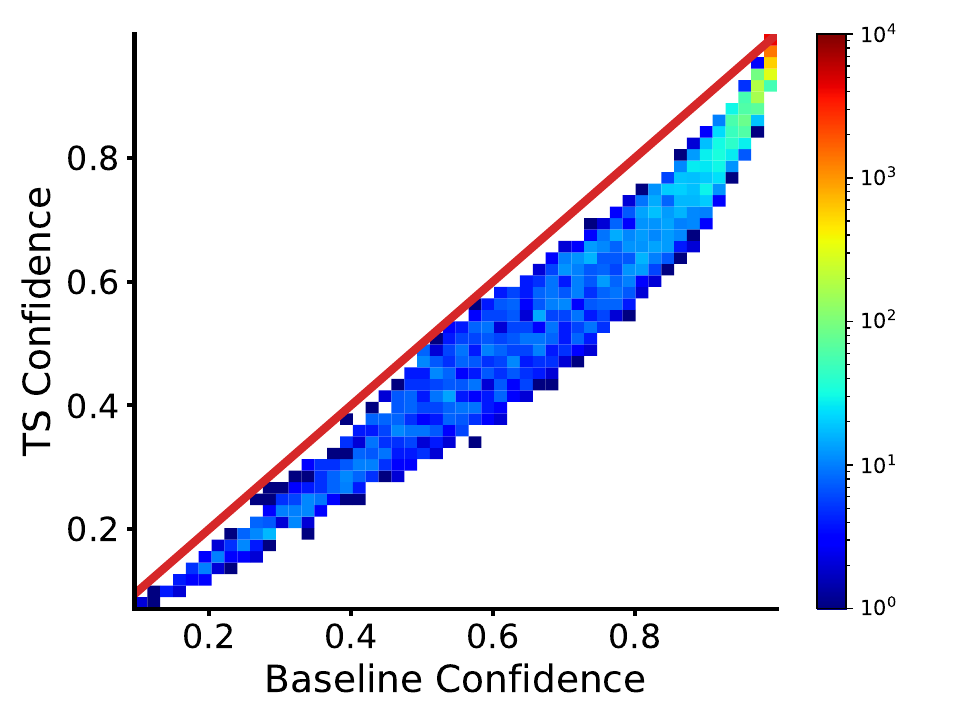}

\caption{Two-dimensional histograms of ResNet-152 confidences on CIFAR-100 baseline confidence (horizontal) vs method confidence (vertical); color encodes sample density and the red diagonal marks no change. SRC (left) increases or decreases confidences; TS (right) only maintains or reduces them.  }
\label{figure:confhist}
\end{center}
\end{figure*}

\subsection{SRC Analysis}
To better understand how SRC improves network calibration, we conducted an in-depth analysis of the features and confidence values in the CIFAR-100 model.

\paragraph{Confidence:}
First, we analyzed the change in confidence. Figure \ref{figure:confhist} displays 2-dimensional histograms for the ResNet-152 model trained on CIFAR-100, where the horizontal axis represents each sample's predicted confidence after Baseline training, and the vertical axis represents confidence after SRC (left) or TS (right). The color intensity indicates the number of samples in a given region, with samples along the red diagonal line signifying no change in confidence. 

We found that both SRC and TS maintain a distribution with relatively high number of extremely confidence samples (Red Square in the upper right for both plots in Figure \ref{figure:confhist}). 
A key difference between SRC and TS is that TS strictly maintains or reduces prediction confidence (Figure \ref{figure:confhist}, right, where all samples lie on or below the red line), which is expected given the approach's mathematical formulation. 
This highlights that while TS enhances calibration, it does so in a limited manner - primarily by warping the output distribution, as is common with many post-hoc methods, rather than embedding a more robust representation of certainty and uncertainty within the model weights.

In contrast, SRC modifies the network weights, uniquely enabling both increases and decreases in predicted confidence (Figure \ref{figure:confhist}, left, where samples are distributed above and below the red line). This embeds the notion of calibration into the model's weights rather than masking the confidence problem through a non-learnable transformation.

\paragraph{Feature Representations:}

Next, we conducted a feature analysis on the CIFAR-100 trained model to better understand how modifications to internal representations affect confidence. For this, we examined activation features from the first two layers of the FF head. We compared SRC to LS, as both methods improve confidence through weight modification (TS leaves feature representations unmodified).

\begin{figure}[!ht]
\begin{center}
\centerline{\includegraphics[width=0.85\columnwidth]{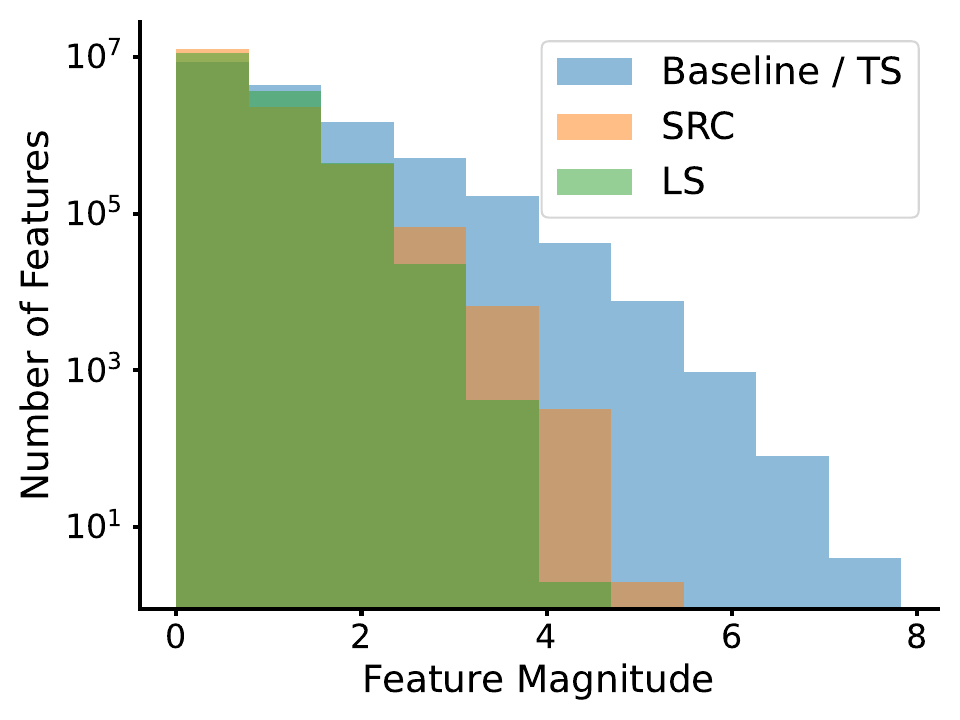}}
\caption{Histograms of FF feature magnitudes for ResNet-152 on CIFAR-100: Baseline/TS (blue), SRC (orange), and LS (green). The Baseline model shows the widest distribution (0–8), LS is more constrained (0–4.5), and SRC shifts the Baseline distribution to a lower maximum (5.5), closely matching LS. This suggests SRC aligns feature representations with the well-calibrated LS model without full retraining.}
\label{figure:featureHist}
\end{center}
\end{figure}

Figure \ref{figure:featureHist} presents histograms of the FF feature magnitudes. The Baseline model (Figure \ref{figure:featureHist}, blue bars) exhibited the widest feature distribution, ranging from 0 to 8, indicating a high degree of feature variability. In contrast, the LS model produced a much more constrained distribution of feature magnitudes, with a maximum around 4.5 (green bars). Similarly, SRC resulted in 
a narrower feature distribution (orange bars) with a maximum of approximately 5.5. 
This demonstrates SRC modifies network weights to achieve similar feature representations to well calibrated retrained models. The benefit of SRC is it imparts these weight modifications in a post-hoc fashion, without the need for complete retraining.

\begin{figure}[!ht]
\begin{center}
\centerline{\includegraphics[width=0.85\columnwidth]{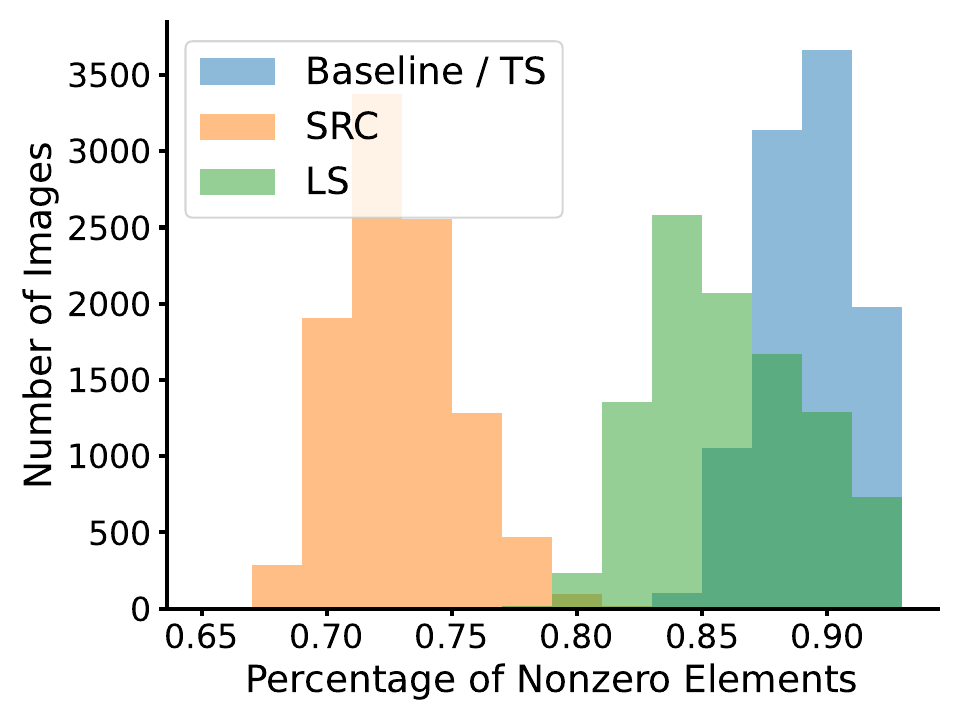}}

\caption{Distributions of nonzero FF layer elements over test samples in ResNet-152 on CIFAR-100 for Baseline/TS (blue), SRC (orange), and LS (green). Baseline representations are densest (82–90\% nonzero) and LS is moderately sparser with partial overlap. SRC achieves substantially sparser representations (67–81\%).}
\label{figure:sparseHist}
\end{center}
\end{figure}

In Figure \ref{figure:sparseHist}, we compare the sparsity of feature representations in the FF layers for each sample across the test dataset. The Baseline model produced the densest representations, with 82\% to 90\% of feature values being nonzero for most samples (Figure \ref{figure:sparseHist}, blue bars). Although the LS model introduced some sparsity, its distribution still overlapped significantly with that of the Baseline (green bars). Notably, applying SRC resulted in significantly sparser representations, with nonzero feature percentages ranging from 67\% to 81\% (orange bars). This increase in sparsity suggests that SRC promotes more compact and efficient feature representations, which may contribute to improved network confidence and calibration.



\paragraph{Synaptic Changes:}

\begin{figure}[!ht]
\centering
\includegraphics[width=0.8\columnwidth]{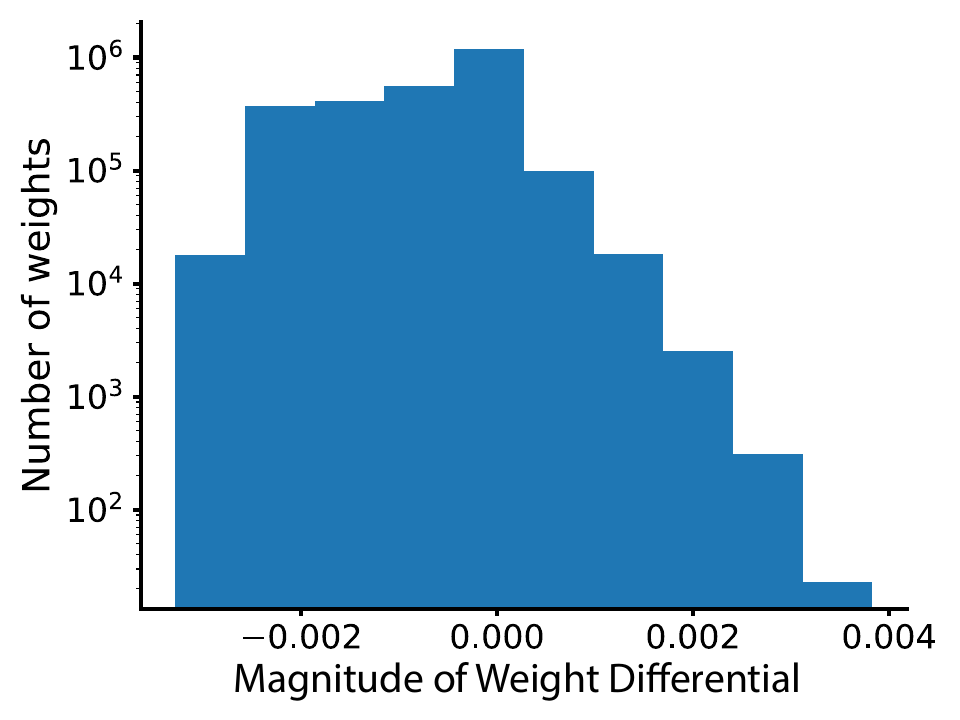}
\caption{Weight changes from SRC for ResNet-152 on CIFAR-100. Most weights decreased, leading to smaller feature magnitudes (similar to LS) and sparser representations.}
\label{figure:srcWeightDiffDist}
\end{figure}

Analysis of the network weights revealed a predominant decrease in weight values after SRC (Figure \ref{figure:srcWeightDiffDist}). This reduction in weight strength may contribute to the narrower distribution of feature magnitudes
and promote sparser representations. Importantly, SRC uniquely modifies network weights post-training, unlike other post-hoc methods that leave them unchanged.


\paragraph{Observed Model Changes Impact on Calibration:}
The observed reduction in synaptic strength suggests that many postsynaptic spikes occur without corresponding presynaptic activity. Such mismatches imply that the associated presynaptic features may be irrelevant, prompting a weight decrease under our Hebbian learning rules. As these weights weaken, the postsynaptic unit becomes less influenced by irrelevant inputs, effectively sharpening the signal-to-noise ratio and enhancing feature selectivity.

Intuitively, lowering feature magnitudes and increasing sparsity encourages each class score to rely on a few strong, selective signals rather than the accumulation of many weak ones. This keeps logit differences moderate on ambiguous or out-of-distribution inputs, preventing the softmax from producing overly confident predictions. The reduced effective capacity also limits the model’s tendency to overfit noise, particularly in low-signal regions. Together, these changes help align predicted confidence more closely with actual accuracy, thereby improving calibration.

\subsection{Scope and Limitations}

A comprehensive theoretical account of how replay combined with local plasticity reshapes deep networks remains an open problem. In this work, we focus on empirical characterization. We found, that across architectures, replay-driven plasticity consistently weakens ineffective pathways and promotes sparser, more selective representations, changes that correlate with improved confidence calibration and sharper class separation. Unlike most calibration approaches that operate purely at the output level, SRC induces interpretable network-level changes through synaptic modification.

SRC differs fundamentally from approaches such as temperature scaling or label smoothing. Rather than applying a top-down transformation to logits, SRC operates bottom-up via local plasticity, reshaping internal representations before they are mapped to class probabilities. This points to a distinct calibration mechanism that directly alters the learned decision structure, motivating future work on a formal theoretical framework.

Although SRC is not as lightweight as scalar post-hoc methods like TS, it offers practical advantages. SRC incurs a one-time, offline cost consisting of a single unsupervised replay phase applied to the feedforward classification head. After replay, inference proceeds with the original network, without additional computation or architectural changes. In contrast, TS applies a temperature transform at every inference step, introducing persistent per-query overhead. Thus, SRC trades a modest one-time cost for zero deployment overhead, which may be preferable when inference efficiency or architectural simplicity is critical.

In this study, SRC was applied only to the feedforward classification head, where calibration is most directly determined. Prior work shows that extending SRC into convolutional layers can further improve robustness and generalization \cite{Delanois_ICMLA23}. We also find that SRC is most stable in models with sufficiently deep classification heads, as shallow heads are more sensitive to local updates that directly perturb logits. Accordingly, our experiments focus on common CNN backbones, including ResNet, VGG, AlexNet, and GoogLeNet. More broadly, SRC is not limited to CNNs: because it operates on feedforward components, it can in principle be applied to other architectures, including transformer MLP blocks.


\section{Discussion}

In this study, we applied an unsupervised Sleep Replay Consolidation (SRC) algorithm to improve calibration - i.e., the alignment between network confidence (the probability mass assigned to a prediction) and accuracy (the likelihood of correctness) - in artificial neural networks (ANNs). We evaluated SRC on several canonical architectures - ResNet-152, VGG, AlexNet, and GoogLeNet - using benchmark datasets including ImageNet and CIFAR-100. We found that SRC significantly improves multiple calibration metrics, including Expected Calibration Error (ECE), leading to a more precise and balanced relationship between confidence and accuracy. SRC matched or outperformed competing methods such as Temperature Scaling and Label Smoothing. Crucially, SRC combines the advantages of post-hoc methods (applicable to fully trained models) and retraining approaches (modifying model weights), while mitigating their drawbacks by avoiding oversimplified output remapping and the substantial computational cost of retraining.

Since its discovery in the 1950s \cite{Aserinsky1953}, Rapid Eye Movement (REM) sleep has remained enigmatic. Although REM sleep is highly conserved across species \cite{Peever2017}, follows a stereotyped developmental trajectory \cite{Blumberg2020}, comprises ~20 of human sleep \cite{Carskadon2011}, and is associated with multiple cognitive functions \cite{Stickgold2000,Mednick2003}, its functional necessity remains unclear; notably, humans can survive without REM sleep \cite{Siegel2001}. Recent advances have opened new avenues for probing REM sleep mechanisms. Two findings are particularly relevant here: (a) evidence for memory-specific synaptic weakening and pruning during REM sleep \cite{Yang2014,Zhou2020}, and (b) a shift in the excitatory/inhibitory balance toward inhibition \cite{Tamaki2020}. Together, these results suggest that REM sleep may improve memory representations by increasing sparseness and reducing overlap between memory traces. Consistent with these observations, our mechanistic analysis shows that SRC predominantly drives synapses toward more negative values, enhancing cross-inhibition, sharpening memory selectivity, and reducing interference.

In humans, sleep strengthens the coupling between confidence and correctness, and even a short pre-retrieval nap improves confidence calibration while reducing high-confidence false identifications \cite{bonilla2025nap}. This behavioral effect aligns with a neural view in which confidence is intrinsic to circuit dynamics rather than a post hoc report. At the single-neuron level, primate parietal neurons jointly encode decisions and their certainty \cite{kiani2009confidence}, and rodent orbitofrontal neurons predict confidence across sensory modalities \cite{masset2020confidence}. Together, these findings motivate sleep-inspired calibration in artificial networks: if biological systems improve confidence reliability through offline consolidation that reshapes internal representations, then a post-training replay phase that updates network weights can analogously refine probabilistic confidence without labeled retraining.

SRC reframes ANN calibration as a structural reorganization problem by introducing an unsupervised “sleep” phase in which the model self-rehearses using noise sampled from learned input statistics. Local learning rules reinforce reliable patterns and weaken spurious ones, aligning confidence with robust feature co-occurrence. Confidence is expressed only when a full evidence pattern consistently activates a class, reducing overconfidence and improving uncertainty estimation. Unlike temperature scaling, which globally rescales logits, or label smoothing, which uniformly penalizes confident predictions, SRC reshapes internal representations by pruning unreliable connections and increasing feature sparsity. Extending this approach may enable improved confidence estimation in larger models, including LLMs, which remain poorly calibrated \cite{zhu2023calibration,lyu2025calibrating}.

In conclusion, we introduced an unsupervised SRC algorithm inspired by the biological role of sleep to improve calibration in artificial neural networks. SRC is the only post-hoc method considered here that directly modifies network weights, allowing confidence to both increase and decrease through replay-driven plasticity rather than global output smoothing. By leveraging noisy reactivation and Hebbian-like learning, SRC achieves robust improvements across calibration metrics while preserving the deployment advantages of post-hoc methods and avoiding retraining costs. More broadly, our results establish a new class of biologically inspired post-hoc calibration methods based on replay-driven weight adaptation, opening the door to extensions incorporating multi-phasic sleep dynamics, richer replay structure, and more sophisticated plasticity rules.

\section*{Impact Statement}



Model calibration is a key component of reliability across tasks, reflecting the alignment between a model’s predicted confidence and its true likelihood of being correct. Poor calibration can lead to critical failures, either by over-trusting inaccurate predictions or under-utilizing reliable ones. While the biological basis of human confidence is not fully understood, sleep has been shown to influence not only memory consolidation but also the alignment between confidence and performance. 
Thus, our results may open a new avenue for developing biologically inspired calibration methods.

Ethics is a fundamentally human trait, raising the question of whether safe and ethical AI can be achieved without introducing human-like mechanisms. While explicit rules can be imposed, no fixed rule set can cover the wide range of safety- and ethics-critical situations an AI may encounter, pointing to a deeper issue: which features of the human brain are essential for developing notions of safety and ethics? Sleep is a fundamental property of biological intelligence, during which recent experiences are replayed to support long-term memory formation, generalization beyond episodic experience, and the emergence of new insights. An open question is whether sleep-like replay can similarly support the emergence of more natural, ethics-related behaviors in AI - a possibility that may be essential for building systems that are not only capable, but genuinely safe and ethically aligned.

\section*{Acknowledgements}
This work was supported by NIH (1R01MH125557 and 1RFNS132913), NSF (EFMA-2223839).

\bibliography{icml2026}
\bibliographystyle{icml2026}

\newpage
\appendix
\onecolumn
\section{Appendix}

\subsection{Metrics} \label{supMetrics}
\subsubsection{Expected Calibration Error (ECE)}
\begin{equation}
\text{ECE} = \sum_{m=1}^{M} \frac{|B_m|}{N} \left| \text{acc}(B_m) - \text{conf}(B_m) \right| \\
\end{equation}


Where:
\begin{itemize}
    \item \( N \) is the total number of samples.
    \item \( M \) is the total number of bins.
    \item \( B_m \) is the set of samples whose predicted confidence falls within the \( m \)th bin.
    \item \( |B_m| \) is the number of samples in bin \( B_m \).
    \item \( \text{acc}(B_m) \) is the accuracy of samples in bin \( B_m \), defined as:
    \begin{equation}
        \text{acc}(B_m) = \frac{1}{|B_m|} \sum_{i \in B_m} \mathbf{1} (\hat{y}_i = y_i),
    \end{equation}
    where \( \hat{y}_i \) is the predicted class, \( y_i \) is the true class, and \( \mathbf{1} (\cdot) \) is the indicator function.
    \item \( \text{conf}(B_m) \) is the average confidence of samples in bin \( B_m \), given by:
    \begin{equation}
        \text{conf}(B_m) = \frac{1}{|B_m|} \sum_{i \in B_m} \hat{p}_i,
    \end{equation}
    where \( \hat{p}_i \) is the model's predicted probability for the correct class \( \hat{y}_i \).
\end{itemize}

\subsubsection{Negative Log-Likelihood (NLL)}
\begin{equation}
\text{NLL} = - \sum_{i=1}^{N} \log (\hat{p}_i)
\end{equation}

Where:

\begin{itemize}
\item \( N \) is the number of samples,
\item \( \hat{p}_i \) is the model's predicted probability for the correct class 
\end{itemize}

\subsubsection{Brier Score}



\[
\text{Brier Score} = \frac{1}{N} \sum_{i=1}^{N} \sum_{c=1}^{C} (p_{ic} - \mathbf{1}(y_i = c))^2
\]

Where:

\begin{itemize}
\item \( N \) is the number of samples,
\item \( C \) is the number of classes,
\item \( p_{ic} \) is the predicted probability of the \( i \)-th sample belonging to class \( c \),
\item \( \mathbf{1}(y_i = c) \) is the indicator function, which equals 1 if the true label \( y_i \) of the \( i \)-th sample is class \( c \), and 0 otherwise.
\end{itemize}

\subsubsection{Entropy}
\begin{equation}
\text{Entropy} = -\frac{1}{N} \sum_{i=1}^{N} \sum_{c}^{C} p_{ic} \log p_{ic}
\end{equation}
Where:

\begin{itemize}
\item \( N \) is the number of samples,
\item \( C \) is the number of classes,
\item \( p_{ic} \) is the predicted probability of the \( i \)-th sample belonging to class \( c \),
\end{itemize}

\subsection{Standard Deviations}
Table \ref{apptab:std} is a table of standard deviations across 10 trials.

\subsection{Sleep Hyperparameters}
Table \ref{apptab:hyper} is a table of sleep hyperparameters.

\begin{table*}[t!]
\centering
\resizebox{0.8\textwidth}{!}{%
\begin{tabular}{@{}cccccc@{}}
\toprule
 & Accuracy & ECE & NLL & Brier & Entropy \\
\midrule
CIFAR 100\\
baseline & 0.0000 & 0.000000 & 0.000000 & 0.000000 & 0.000000 \\
baseline + SRC & 0.0583 & 0.000928 & 0.000575 & 0.026980 & 0.008026 \\
baseline + TS & 0.0000 & 0.000003 & 0.000000 & 0.000178 & 0.000033 \\
baseline + SRC + TS & 0.0583 & 0.000692 & 0.000387 & 0.004049 & 0.000727 \\
baseline + LS & 0.1078 & 0.002367 & 0.005212 & 0.016566 & 0.019289 \\
baseline + Focal  & 0.1664 & 0.003875 & 0.007602 & 0.584977 & 0.029837 \\
\\
Imagenet\\
baseline | alexnet & 0.0000 & 0.000000 & 0.000000 & 0.000001 & 0.000000 \\
baseline + SRC | alexnet & 0.0198 & 0.000345 & 0.000140 & 0.030474 & 0.004528 \\
baseline + TS | alexnet & 0.0000 & 0.000001 & 0.000000 & 0.000001 & 0.000000 \\
baseline + SRC + TS | alexnet & 0.0203 & 0.000319 & 0.000207 & 0.003985 & 0.000394 \\
\\
baseline | vgg19 & 0.0000 & 0.000000 & 0.000000 & 0.000001 & 0.000000 \\
baseline + SRC | vgg19 & 0.0020 & 0.000039 & 0.000002 & 0.000771 & 0.000106 \\
baseline + TS | vgg19 & 0.0000 & 0.000007 & 0.000000 & 0.000002 & 0.000000 \\
baseline + SRC + TS | vgg19 & 0.0020 & 0.000017 & 0.000002 & 0.000181 & 0.000005 \\
\\
baseline | resnet50 & 0.0000 & 0.000000 & 0.000000 & 0.000000 & 0.000000 \\
baseline + SRC | resnet50 & 0.0122 & 0.000189 & 0.000110 & 0.001717 & 0.000149 \\
baseline + TS | resnet50 & 0.0000 & 0.000000 & 0.000000 & 0.000001 & 0.000000 \\
baseline + SRC + TS | resnet50 & 0.0122 & 0.000282 & 0.000089 & 0.002088 & 0.000278 \\
\\
baseline | resnet152 & 0.0000 & 0.000000 & 0.000000 & 0.000001 & 0.000000 \\
baseline + SRC | resnet152 & 0.0165 & 0.000213 & 0.000247 & 0.006651 & 0.000285 \\
baseline + TS | resnet152 & 0.0000 & 0.000000 & 0.000000 & 0.000001 & 0.000001 \\
baseline + SRC + TS | resnet152 & 0.0189 & 0.000286 & 0.000226 & 0.005213 & 0.000434 \\
\\
baseline | resnet152FF & 0.0000 & 0.000000 & 0.000000 & 0.000001 & 0.000000 \\
baseline + SRC | resnet152FF & 0.1051 & 0.001920 & 0.004277 & 0.087227 & 0.034782 \\
baseline + TS | resnet152FF & 0.0000 & 0.000020 & 0.000000 & 0.000627 & 0.000196 \\
baseline + SRC + TS | resnet152FF & 0.1081 & 0.001579 & 0.004096 & 0.046032 & 0.007520 \\
\\
baseline | googlenet & 0.0000 & 0.000000 & 0.000000 & 0.000000 & 0.000001 \\
baseline + SRC | googlenet & 0.0924 & 0.001064 & 0.011851 & 0.173751 & 0.004951 \\
baseline + TS | googlenet & 0.0000 & 0.000012 & 0.000000 & 0.000001 & 0.000001 \\
baseline + SRC + TS | googlenet & 0.0944 & 0.000860 & 0.012037 & 0.134106 & 0.017651 \\
\\
baseline | googlenetFF & 0.0000 & 0.000000 & 0.000000 & 0.000001 & 0.000001 \\
baseline + SRC | googlenetFF & 0.0776 & 0.003493 & 0.004056 & 0.194901 & 0.106859 \\
baseline + TS | googlenetFF & 0.0000 & 0.000000 & 0.000001 & 0.000003 & 0.000001 \\
baseline + SRC + TS | googlenetFF & 0.0796 & 0.001139 & 0.001607 & 0.034019 & 0.002915 \\
\bottomrule
\end{tabular}%
}
\caption{Standard deviations across 10 trials}
\label{apptab:std}
\end{table*}

\begin{algorithm}[t]
\caption{Sleep Replay Consolidation (SRC)}
\label{alg:src}
\begin{algorithmic}[1]
\STATE \textbf{Input:} network $nn$, input statistics $I$, scaling factors $scales$, thresholds $thresholds$
\STATE Initialize membrane voltages $v \leftarrow 0$ for all neurons
\FOR{$t = 1$ to $T_s$}
    \STATE $S \leftarrow 0$
    \STATE Convert input $I$ to Poisson-distributed spiking activity
    \FOR{$l = 2$ to $n$}
        \STATE $\alpha \leftarrow scales(l-1)$, $\beta \leftarrow thresholds(l)$
        \STATE $v^{(l)} \leftarrow \lambda v^{(l)} + \alpha \, W^{(l,l-1)} S^{(l-1)}$
        \STATE Set $S_i^{(l)} \leftarrow 1$ where $v_i^{(l)} > \beta$
        \STATE Reset $v_i^{(l)} \leftarrow 0$ where $v_i^{(l)} > \beta$
    \ENDFOR
    \FOR{$l = 2$ to $n$}
        \FOR{all synapses $(i,j)$}
            \IF{$S_j^{(l)} = 1$ and $S_i^{(l-1)} = 1$}
                \STATE $W_{i,j}^{(l,l-1)} \leftarrow W_{i,j}^{(l,l-1)} + inc$
            \ELSIF{$S_j^{(l)} = 1$ and $S_i^{(l-1)} = 0$}
                \STATE $W_{i,j}^{(l,l-1)} \leftarrow W_{i,j}^{(l,l-1)} - dec$
            \ENDIF
        \ENDFOR
    \ENDFOR
\ENDFOR
\STATE \textbf{return} $W$
\end{algorithmic}
\end{algorithm}


\begin{table}[ht]
  \centering
  \resizebox{\textwidth}{!}{%
  \begin{tabular}{lccccccc}
    \toprule
    Simulation & Time Steps & dt & Decay Rate & Max Spiking RateRate & Postive STDP & Negative STDP & Spiking Thresholds \\ 
    \midrule
    \addlinespace
    ResNet 152 CIFAR 100 SRC & 443 & 0.001 & 0.9521040578368124 & 299.64109973100756 & 0.000716159191361385 & -0.00046873584104501213 & [18.08956309358943, 20.4728432017464, 17.0355215370257] \\ 
    \addlinespace
    ResNet 152 CIFAR 100 SRC TS & 443 & 0.001 & 0.9521040578368124 & 299.64109973100756 & 0.000716159191361385 & -0.00046873584104501213 & [18.08956309358943, 20.4728432017464, 17.0355215370257] \\ 
    \addlinespace
    ImageNet AlexNet SRC & 492 & 0.001 & 0.9836698145272654 & 96.77346924873888 & 0.00017228598853676435 & -0.0003060770957571102 & [0.3969962502366198, 19.801537664023588, 17.8318793443642] \\ 
    \addlinespace
    ImageNet AlexNet SRC TS & 492 & 0.001 & 0.9836698145272654 & 96.77346924873888 & 0.00017228598853676435 & -0.0003060770957571102 & [0.3969962502366198, 19.801537664023588, 17.8318793443642] \\ 
    \addlinespace
    ImageNet GoogLeNet SRC & 297 & 0.001 & 0.9230026016980454 & 57.69634835829613 & 4.994787627593506e-05 & -0.0009992231413448339 & [1.061303932245, 1.0, 1.0] \\ 
    \addlinespace
    ImageNet GoogLeNet SRC TS & 297 & 0.001 & 0.9230026016980454 & 57.69634835829613 & 4.994787627593506e-05 & -0.0009992231413448339 & [1.061303932245, 1.0, 1.0] \\ 
    \addlinespace
    ImageNet GoogLeNet FF SRC & 216 & 0.001 & 0.9190067682396723 & 373.602370340872 & 0.0005747601430541048 & -0.0006767870342010161 & [23.167565098799404, 1.0, 1.0] \\ 
    \addlinespace
    ImageNet GoogLeNet FF SRC TS & 216 & 0.001 & 0.9190067682396723 & 373.602370340872 & 0.0005747601430541048 & -0.0006767870342010161 & [23.167565098799404, 1.0, 1.0] \\ 
    \addlinespace
    ImageNet Resnet 152 SRC & 142 & 0.001 & 0.9967226847955232 & 270.3719637375567 & 0.0006835896002156544 & -8.554105136507242e-05 & [5.960976752777733, 1.0, 1.0] \\ 
    \addlinespace
    ImageNet Resnet 152 SRC TS & 142 & 0.001 & 0.9967226847955232 & 270.3719637375567 & 0.0006835896002156544 & -8.554105136507242e-05 & [5.960976752777733, 1.0, 1.0] \\ 
    \addlinespace
    ImageNet Resnet 152 FF SRC & 133 & 0.001 & 0.9541199194080845 & 168.0334048240091 & 0.0004137546177176504 & -0.0003593967239143372 & [24.44266615208508, 7.036235824740411, 0.29729907661881144] \\ 
    \addlinespace
    ImageNet Resnet 152 FF SRC TS & 133 & 0.001 & 0.9541199194080845 & 168.0334048240091 & 0.0004137546177176504 & -0.0003593967239143372 & [24.44266615208508, 7.036235824740411, 0.29729907661881144] \\ 
    \addlinespace
    ImageNet Resnet 50 SRC & 191 & 0.001 & 0.9910941324849671 & 206.8851715685009 & 0.0008062400912877635 & -0.00028408173096335684 & [9.083349761366986, 0.843135195703389, 9.326052313371362] \\ 
    \addlinespace
    ImageNet Resnet 50 SRC TS & 191 & 0.001 & 0.9910941324849671 & 206.8851715685009 & 0.0008062400912877635 & -0.00028408173096335684 & [9.083349761366986, 0.843135195703389, 9.326052313371362] \\ 
    \addlinespace
    ImageNet VGG 19 SRC & 375 & 0.001 & 0.9977038423803284 & 103.20411158519192 & 0.0004419052103542043 & -4.595885712289623e-05 & [6.777582201627136, 16.049922603940825, 19.43243751069473] \\ 
    \addlinespace
    ImageNet VGG 19 SRC TS & 375 & 0.001 & 0.9977038423803284 & 103.20411158519192 & 0.0004419052103542043 & -4.595885712289623e-05 & [6.777582201627136, 16.049922603940825, 19.43243751069473] \\ 
    \addlinespace
    \bottomrule
  \end{tabular}
  }
  \caption{Sleep hyperparameters for each simulation.}
  \label{apptab:hyper}
\end{table}




\end{document}